\documentclass{article}

\usepackage{microtype}
\usepackage{graphicx}
\usepackage{subcaption}
\usepackage{booktabs} 

\usepackage{hyperref}

\usepackage[preprint]{icml2026}

\usepackage{amsmath}
\usepackage{amssymb}
\usepackage{mathtools}
\usepackage{amsthm}

\usepackage[capitalize,noabbrev]{cleveref}

\theoremstyle{plain}
\newtheorem{theorem}{Theorem}[section]
\newtheorem{proposition}[theorem]{Proposition}

\theoremstyle{definition}
\newtheorem{definition}[theorem]{Definition}

\theoremstyle{remark}

\usepackage[textsize=tiny]{todonotes}

\usepackage{amssymb}
\usepackage{pifont}

\newcommand{\norm}[1]{{\left\Vert #1 \right\Vert }}
\newcommand{\innp}[1]{{\langle #1 \rangle }}

\usepackage{mathtools}
\usepackage{bbm}
\usepackage{csquotes}
\usepackage{graphicx}
\usepackage{subcaption}
\usepackage{booktabs}
\usepackage{mwe}
\usepackage{wrapfig}
\usepackage{comment}
\usepackage{amsthm}
\usepackage{multirow}

\newcommand{\R}{\mathbb R}

\newcommand{\set}[1]{{\{ #1 \}}}

\newcommand{\defi}{\stackrel{\mathrm{\scriptscriptstyle def}}{=}}

\usepackage[acronym]{glossaries}
\glsdisablehyper

\usepackage{titletoc}
\usepackage{placeins}

\newacronym{greedyaf}{SparseSwaps}{SparseSwaps}
\newacronym{imp}{IMP}{Iterative Magnitude Pruning}
\newacronym{per}{PERP}{Parameter-Efficient Retraining after Pruning}
\newacronym{ft}{FT}{Fine-Tuning}
\newacronym{lora}{LoRA}{Low-Rank Adaptation}
\newacronym{llm}{LLM}{Large Language Model}
\newacronym{gpt}{GPT}{Generative Pretrained Transformer}
\newacronym{nlp}{NLP}{Natural Language Processing}
\newacronym{bn}{BN}{Batch-Normalization}
\newacronym{ln}{LN}{Layer-Normalization}
\newacronym{ip}{IP}{Integer Programming}
\newacronym{svd}{SVD}{Singular Value Decomposition}
\newacronym{dsnot}{DSnoT}{DSnoT}

\newcommand{\glsshort}[1]{\glsentryshort{#1}}

\icmltitlerunning{SparseSwaps: Tractable LLM Pruning Mask Refinement at Scale}

\begin{document}

\twocolumn[
  \icmltitle{SparseSwaps: Tractable LLM Pruning Mask Refinement at Scale}



  \icmlsetsymbol{equal}{*}

  \begin{icmlauthorlist}
    \icmlauthor{Max Zimmer}{zib}
    \icmlauthor{Christophe Roux}{zib}
    \icmlauthor{Moritz Wagner}{zib}
    \icmlauthor{Deborah Hendrych}{zib}
    \icmlauthor{Sebastian Pokutta}{zib}
  \end{icmlauthorlist}

  \icmlaffiliation{zib}{Department for AI in Society, Science, and Technology, Zuse Institute Berlin, Germany\\
  Institute of Mathematics, Technische Universität Berlin, Germany}

  \icmlcorrespondingauthor{Max Zimmer}{zimmer@zib.de}

  \icmlkeywords{Machine Learning, ICML}

  \vskip 0.3in
]



\printAffiliationsAndNotice{}  

\begin{abstract}
  The resource requirements of neural networks can be significantly reduced through pruning -- the removal of seemingly less important parameters. However, for LLMs, full retraining to recover pruning-induced performance degradation is often prohibitive and classical approaches such as magnitude pruning are suboptimal on Transformers. State-of-the-art methods hence solve a layer-wise \emph{mask selection problem}: finding a pruning mask that minimizes per-layer pruning error on a small set of calibration data. Exactly solving this problem is computationally infeasible due to its combinatorial nature and the size of the search space, and existing approaches rely on approximations or heuristics. We demonstrate that the mask selection problem can be made drastically more tractable at LLM scale. To that end, we decouple the rows by enforcing equal sparsity levels per row. This allows us to derive optimal \emph{1-swaps} (exchanging one kept and one pruned weight) computable efficiently via the Gram matrix. We propose a simple 1-swap algorithm that warmstarts from any pruning mask, runs efficiently on GPUs at LLM scale, and is essentially hyperparameter-free. Our approach reduces per-layer pruning error by up to 60\% over Wanda \citep{Sun2023} and consistently improves perplexity and zero-shot accuracy across state-of-the-art GPT architectures.
\end{abstract}

\section{Introduction}
\emph{Pruning after training} \citep{Han2015, Gale2019, Lin2020, Hoefler2021, Zimmer2022} is a state-of-the-art technique to reduce the resource requirements of neural networks. A simple yet effective approach to obtain such \emph{sparse} models starts from a pretrained \emph{dense} model, removes seemingly unimportant parameters based on their magnitude, and requires retraining to compensate for pruning-induced performance degradation. However, while the inexpensive, data-free magnitude criterion has often achieved strong performance on traditional architectures \citep{Gale2019, Zimmer2021}, pruning has undergone a paradigm shift with the rise of pretrained foundation models such as \glspl{llm}.

First, the size of the models has shifted the focus toward retraining-free pruning criteria, as retraining is often computationally expensive if not infeasible, with parameter-efficient fine-tuning \citep{Lialin2023a, Zimmer2023a} being an exception. Secondly, systematic activation outliers \citep{Dettmers2022} and highly important \emph{super-weights} \citep{yuSuperWeightLarge2025} in sufficiently large \emph{Transformers} \citep{Vaswani2017} have rendered magnitude pruning no better than random pruning for \glspl{llm} \citep{Sun2023, Yin2023a}. Lastly, state-of-the-art methods \citep{Frantar2023a, Sun2023, Zhang2024} prune \emph{layer-wise}: they split the pruning problem into per-layer subproblems, pruning layers sequentially and independently using a small calibration dataset to estimate parameter importance. Rather than optimizing the \emph{global} loss, such approaches minimize a per-layer \emph{local} pruning loss. Specifically, for a single layer with calibration input matrix $X\in \R^{d_{in} \times B}$ and weights $W\in \R^{d_{out} \times d_{in}}$, the objective becomes
\begin{equation}\label{eq:mask_selection_problem}
    \min_{M}\norm{W X - (M \odot W) X}_F^2,
\end{equation}
where $M \in \set{0,1}^{d_{out} \times d_{in}}$ is a binary pruning mask achieving a desired level of sparsity, e.g., $\norm{M}_0 \leq k$ for unstructured sparsity, and $\odot$ denotes the element-wise multiplication or Hadamard product. Here, $B = N \cdot L$ with $N$ being the number of samples in the calibration batch and $L$ being the sequence length.

Solving this combinatorial \emph{mask selection problem} to optimality is NP-hard due to feature correlations: selecting $k$ of $d_{out}\cdot d_{in}$ weights yields a cardinality-constrained binary quadratic program (a best-subset selection variant). Even for a single row $i$ the problem remains hard, despite reducing to
\begin{equation*}
   \min_{m_i}\norm{w_i^{\top} X - (m_i \odot w_i)^{\top} X}_2^2
 \end{equation*}
 where $w_i \in \R^{d_{in}}$ and $m_i \in \set{0,1}^{d_{in}}$ denote the $i$-th row of $W$ and $M$, respectively. While \gls{ip} solvers could theoretically provide optimal solutions, the combinatorial search over mask entries makes this infeasible for \glspl{llm}. In practice, existing methods therefore relax \autoref{eq:mask_selection_problem} or approximate it.

However, with deployed \glspl{llm} now serving millions of users, it becomes increasingly worthwhile to invest substantial resources to obtain pruned models that reach high performance, because the pruning cost is paid once during training whereas inference costs scale with the number of requests. In this work, we revisit the per-layer mask selection problem and demonstrate that it can be operationalized at \gls{llm} scale, enabling monotone improvements with each optimization step rather than relying on proxy importance scores. To that end, we observe that enforcing equal sparsity levels across rows ensures \emph{row-wise separability} that yields independent objectives. This makes the problem drastically more tractable and still leads to good practical performance for \glspl{llm}. Instead of trying to obtain exact solutions via \gls{ip} solvers, we instead propose a GPU-accelerated local optimization algorithm based on \emph{1-swaps} (exchanging one kept and one pruned weight) that perform \emph{exact and efficient local refinement with incremental cost updates} using the Gram matrix $G = XX^{\top}$ to monotonically decrease the objective from any warmstart. 

\begin{figure*}[t]
  \centering
  \includegraphics[width=\linewidth]{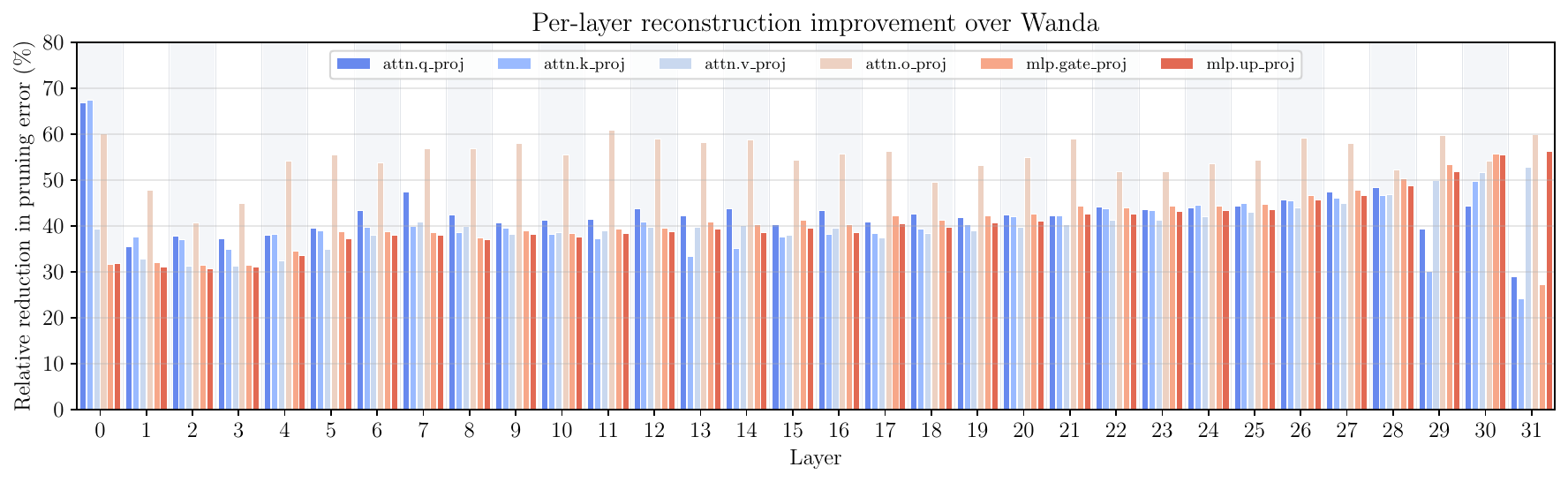}
  \caption{Per-layer relative reduction in local pruning error compared to Wanda. The plot shows results for \textsc{LLaMA-3.1-8B}, 60\% unstructured sparsity and 100 1-swap iterations.}
  \label{fig:layer_error}
\end{figure*}

The resulting method, which we term \acrshort{greedyaf}, can start from any warmstart mask, evaluates the exact per-row quadratic loss, and is scalable, parallelizable across rows, almost hyperparameter-free, and deterministic for a fixed warmstart. With only a few 1-swap iterations, it can reduce the per-layer pruning error by up to 60\% compared to Wanda and improves final perplexity and zero-shot accuracy across architectures. Our approach is a post-hoc refinement of existing pruning methods that can significantly improve upon the state of the art for unstructured, per-row, or $N\text{:}M$ sparsity.
 
 \paragraph{Contributions.} Our contributions are as follows:
 \begin{enumerate}
    \item \textbf{Making the mask selection problem tractable.} We observe that a) enforcing equal sparsity levels per row decouples the rows, and that b) optimal 1-swaps (exchanging one kept and one pruned weight) can be evaluated efficiently using the Gram matrix $G = XX^{\top}$ of the calibration data, ensuring efficient lookups when determining the most beneficial swap.
    \item \textbf{\glsshort{greedyaf}: a practical post-hoc pruning algorithm.} Building on these observations, we propose \glsshort{greedyaf}, a plug-and-play 1-swap refinement that starts from any warmstart mask and monotonically decreases the exact per-row objective under per-row or $N\text{:}M$ constraints. In particular, \glsshort{greedyaf} is almost hyperparameter-free, completely parallelizable across rows and scalable to \glspl{llm}.
    \item \textbf{Computational study.} We verify our hypotheses on state-of-the-art \gls{gpt} architectures and demonstrate that \glsshort{greedyaf} delivers large reductions in local pruning error (up to 60\% per-layer error reduction over Wanda) and strong perplexity and zero-shot gains across a wide range of different \glspl{llm}. Surprisingly, when pruning-induced degradation is low, we find that while \glsshort{greedyaf} still consistently and drastically reduces the local error, these gains do not necessarily translate to perplexity improvements, highlighting inherent limitations of layer-wise mask optimization as an approximation to the global objective. We conduct a series of ablations highlighting the advantages and drawbacks of the proposed approach.
 \end{enumerate}

\paragraph{Further related work.}
\emph{Post-training pruning} has a long history, and while \emph{magnitude pruning} \citep{Janowsky1989, Han2015} is among the most popular criteria, it is not the only one \citep[cf.][]{LeCun1989, Hassibi1992, Molchanov2016, Yeom2019}; see \citet{Hoefler2021} for a comprehensive review. Despite their simplicity, magnitude-based methods have been shown to produce sparse models competitive with far more complex algorithms for convolutional architectures \citep{Gale2019, Zimmer2021}. For \glspl{llm}, however, magnitude pruning is argued to be unsuited \citep{Yin2023a}. Consequently, there is growing interest in criteria beyond magnitude that achieve high performance on \glspl{llm}, and do so without requiring an expensive retraining procedure \citep{Kwon2022, Frantar2023a, Sun2023, Zhang2024}. In this work, we develop a post-hoc refinement of existing methods, rather than proposing a new criterion. A related approach, \glsshort{dsnot} \citep{Zhang2023a}, also performs iterative weight swaps but differs significantly in its optimization strategy. Inspired by \emph{dynamic sparse training} \citep[cf.][]{Evci2019}, \glsshort{dsnot} prunes and regrows weights based on expected reconstruction-error improvements, using feature means and variances as surrogates. While effective, it does not guarantee a monotonic decrease in the true pruning error, whereas our method does. We compare the two empirically and find that \glsshort{greedyaf} consistently outperforms \glsshort{dsnot}.

\emph{Subset selection and \gls{ip} approaches.} To solve \autoref{eq:mask_selection_problem} to global optimality, which can be formulated as a \emph{mixed-integer nonlinear program (MINLP)}, several efficient open-source solvers are available, including SCIP \citep{BolusaniEtal2024OO}, Bonmin \citep{bonami2008algorithmic}, SHOT \citep{lundell2022supporting} and Boscia \citep{2022_HendrychTroppensBesanconPokutta_Convexintegerfrankwolfe}, among others. While we demonstrate how the problem can be made drastically more tractable, explicit solution remains very time-consuming for large instances; we therefore opt for a GPU-friendly 1-swap approach that avoids moving large tensors to the CPU for \gls{ip} solvers. We leave such an extension for future work.

\section{Methodology}\label{sec:methodology}
In the following, we use uppercase letters for matrices ($W$, $X$, $M$) and lowercase letters for scalars and vectors. Matrix entries are denoted $W_{ij}$ for the element in row $i$, column $j$. Rows of matrices are denoted with lowercase subscripts: $w_i$ represents the $i$-th row of matrix $W$. Row and column slices use colon notation: $X_{j,:}$ for the $j$-th row and $X_{:,k}$ for the $k$-th column. We use $\odot$ for element-wise multiplication, $\norm{\cdot}_F$ for the Frobenius norm, and $\norm{\cdot}_2$ for the $\ell_2$ norm.
\subsection{Preliminaries}\label{sec:preliminaries}
Before describing our proposed method, we make several assumptions and observations that make the problem tractable. 

\subsubsection{Row Decoupling through Equal Per-Row Sparsity}
First, note that the objective in \autoref{eq:mask_selection_problem} decomposes into a sum of $d_{out}$ row-wise quadratics,
\begin{equation*}
\norm{(W - M \odot W) X}_F^2 = \sum_{i=1}^{d_{out}} \norm{w_i^{\top} X - (m_i \odot w_i)^{\top} X}_2^2
\end{equation*}
where $w_i \in \R^{d_{in}}$ and $m_i \in \set{0,1}^{d_{in}}$ denote the $i$-th row of $W$ and $M$, respectively. If we enforce equal sparsity levels per row, the rows become fully decoupled and can be treated independently, reducing the problem to
\begin{align}
  \label{eq:mask_selection_problem_single_row}
  \min_{m_i}\norm{w_i^{\top} X - (m_i \odot w_i)^{\top} X}_2^2
\end{align}
for each row $i\in \{1, \ldots, d_{out}\}$. This is naturally the case for per-row sparsity constraints (keep $k$ weights per row) and semi-structured $N\text{:}M$ patterns (prune $M\!-\!N$ per block of $M$ weights). For unstructured sparsity, however, the global cardinality constraint $\norm{M}_0 \leq k$ couples rows. \citet{Sun2023} observe that enforcing row-wise sparsity benefits performance for both Wanda and magnitude pruning on \glspl{llm}. We therefore assume per-row sparsity even in nominally unstructured settings, enabling row decoupling without necessarily harming final performance.

As a side note, we can now easily derive the Wanda criterion by upper bounding the error with Jensen's inequality and noting that positive scaling preserves minima, i.e.,
\begin{align}
  &\min_{m_i}\norm{w_i^{\top} X - (m_i \odot w_i)^{\top} X}_F^2\notag \\
   =&\min_{m_i}\sum_{k=1}^{B} \left(\sum_{j=1}^{d_{in}} (1-m_{ij} ) w_{ij} X_{jk}\right)^2 \label{eq:naive_objective} \\ \leq &\min_{m_i}\sum_{k=1}^{B} \left(\sum_{j=1}^{d_{in}} (1-m_{ij})^2 w_{ij}^2 X_{jk}^2\right)\nonumber\\
  = &\min_{m_i} \sum_{j=1}^{d_{in}} (1-m_{ij})^2 w_{ij}^2 \norm{X _{j,:}}_2^2.\label{eq:wanda_objective}
\end{align}
\autoref{eq:wanda_objective} is solved by pruning entries with the smallest saliency $|w_{ij}| \cdot \norm{X_{j,:}}_2$, i.e., precisely the Wanda criterion. Thus, Wanda optimizes an upper bound to the original problem that ignores within-row interactions, making the combinatorial problem tractable.

\subsubsection{Avoiding Intermediate Value Caching Through the Gram Matrix Formulation}
Naively caching all $B \cdot d_{in}$ intermediate products $w_{ij} X_{jk}$ in \autoref{eq:naive_objective} to evaluate candidate masks is prohibitive. To illustrate the scale, consider a single row of the largest matrix in a \textsc{LLaMA-2-7B} Transformer block: the \texttt{up\_proj} matrix with input dimension $d_{in} = 4096$. With $N = 128$ samples and sequence length $L = 4096$ (so $B = N \cdot L = 524{,}288$), caching all products $w_{ij} X_{jk}$ for that row requires $524{,}288 \times 4096 \approx 2.15$ billion float32 values (about 8.6GB); across all 11,008 rows this totals about 94.6TB.

A straightforward way to circumvent this issue is to consider a single row and derive a compact formulation of the per-row loss through the Gram matrix $G \defi XX^{\top} \in \R^{d_{in} \times d_{in}}$. For notational convenience, we drop the row index $i$ throughout the remainder of this section and write $w \in \R^{d_{in}}$ for the row's weight vector and $m \in \{0,1\}^{d_{in}}$ for its mask. The per-row loss from \autoref{eq:mask_selection_problem_single_row} is
\begin{align*}
L &\defi \norm{w^{\top} X - (m \odot w)^{\top} X}_F^2
= \norm{(w - m \odot w)^{\top} X}_F^2 \\
&= (w - m \odot w)^{\top} G (w - m \odot w).
\end{align*}
Hence, the loss depends on $X$ only through the Gram matrix $G$, which can be accumulated on-the-fly as calibration samples pass through the layer: $G = \sum_{b=1}^{B} X_{:,b} X_{:,b}^{\top}$. Unlike the per-row formulation of \autoref{eq:naive_objective}, which would require caching all $B \cdot d_{in}$ intermediate products $w_j X_{jk}$, we only need to maintain the $d_{in} \times d_{in}$ matrix $G$, which is a reduction from $\mathcal{O}(B \cdot d_{in})$ to $\mathcal{O}(d_{in}^2)$ (with $d_{in}$ typically being much smaller than $B$).

\subsubsection{Efficient 1-Swap Evaluation Through Efficient Cost Lookups and Updates}
While the global mask selection problem is NP-hard, we can still make efficient progress via local search. Starting from any feasible mask $m \in \{0,1\}^{d_{in}}$, the idea is to iteratively perform 1-swaps that exchange one kept and one pruned weight to reduce $L$ while preserving the sparsity level. The key observation is that each candidate swap can be evaluated in $\mathcal{O}(1)$ time using $G$ and an auxiliary \emph{correlation vector} $c$. To that end, let $\mathcal{P} \defi \{j : m_j = 0\}$ denote the set of currently pruned weight indices and analogously $\mathcal{U} \defi \{j : m_j = 1\}$ denote the set of unpruned (kept) weight indices. Letting further $\phi_j \defi X_{j,:}^{\top} \in \R^B$ denote the $j$-th row (or feature vector) of $X$, we can write
\begin{equation*}
(w - m \odot w)^{\top} X = \sum_{j=1}^{d_{in}} (1-m_j) w_j X_{j,:} = \sum_{j \in \mathcal{P}} w_j \phi_j^{\top} = r^{\top},
\end{equation*}
where we define the \emph{reconstruction residual} 
\begin{equation*}
  r \defi \sum_{j \in \mathcal{P}} w_j \phi_j \in \R^B,
\end{equation*}
the total contribution of all pruned weights to the layer output. Hence, clearly, the loss is 
\begin{equation*}
  L = \norm{r}_2^2 = r^{\top} r.
\end{equation*}
We define the \emph{correlation vector} $c \in \R^{d_{in}}$ with entries
\begin{align*}
c_i \defi \innp{\phi_i, r} = \innp{\phi_i, \sum_{j \in \mathcal{P}} w_j \phi_j} &= \sum_{j \in \mathcal{P}} w_j \innp{\phi_i, \phi_j} \\ &= \sum_{j \in \mathcal{P}} w_j G_{ij},
\end{align*}
which measures how each feature $\phi_i$ correlates with the current residual. In vector form, $c = G \cdot ((\mathbbm{1} - m) \odot w)$.

\paragraph{Swap cost formula.} A 1-swap removes index $u \in \mathcal{U}$ from the unpruned set (making it pruned) and adds index $p \in \mathcal{P}$ to the unpruned set (making it unpruned). The new residual is $r' = r + w_u \phi_u - w_p \phi_p$, and the change in loss is
\begin{align*}
\Delta L_{u,p} &\defi \norm{r'}_2^2 - \norm{r}_2^2 = \norm{r + w_u \phi_u - w_p \phi_p}_2^2 - \norm{r}_2^2 \\
&= 2w_u \innp{\phi_u, r} + w_u^2 \norm{\phi_u}_2^2 - 2w_p \innp{\phi_p, r} \\
&\quad + w_p^2 \norm{\phi_p}_2^2 - 2w_u w_p \innp{\phi_u, \phi_p}.
\end{align*}
Using $c_i = \innp{\phi_i, r}$ and that $G_{ij} = \innp{\phi_i, \phi_j}$, this simplifies to
\begin{align}
\Delta L_{u,p} &= 2w_u c_u + w_u^2 G_{uu}\notag \\
 &\quad- 2w_p c_p + w_p^2 G_{pp} - 2w_u w_p G_{up}\label{eq:swap_cost}.
\end{align}
Given the precomputed Gram matrix $G$ and correlation vector $c$, each swap evaluation requires only \emph{scalar lookups}. The theoretical complexity of evaluating all possible swaps therefore equals $\mathcal{O}(|\mathcal{U}| \cdot |\mathcal{P}|)$. By systematically testing all $((d_{in} - |\mathcal{P}|) \cdot |\mathcal{P}|)$ possible 1-swap operations (adding one of $|\mathcal U| = d_{in} - |\mathcal{P}|$ unpruned weights to $\mathcal{P}$, removing one of $|\mathcal{P}|$ pruned weights from $\mathcal{P}$) evaluating the improvement using the above expression, we iteratively pick a best swap and update the mask until we have reached a satisfactory solution or one optimal w.r.t.\ 1-swap operations. The only issue that remains is to update the correlation vector after each swap.

\paragraph{Correlation vector update.} After accepting a swap $(u^*, p^*)$, the residual changes to $r' = r + w_{u^*} \phi_{u^*} - w_{p^*} \phi_{p^*}$. The correlation vector updates as
\begin{equation}\label{eq:c_update}
c_i \leftarrow c_i + w_{u^*} G_{i,u^*} - w_{p^*} G_{i,p^*},
\end{equation}
or in vector form, $c \leftarrow c + w_{u^*} G_{:,u^*} - w_{p^*} G_{:,p^*}$. This only requires accessing two columns of $G$ and costs $\mathcal{O}(d_{in})$.

\paragraph{Why picking $p$ and $u$ separately is suboptimal.} The interaction term $-2w_u w_p G_{up}$ in \autoref{eq:swap_cost} shows that the best $u$ depends on the chosen $p$ (and vice versa). Consequently, selecting $p$ and $u$ based on their individual effects can yield a detrimental swap, as the following example for the scalar case with $B = 1$ and $d_{in} = 4$ shows. Let the current pruned weight contributions be $\{+10, -1\}$, so $r = 9$ and $L = 81$, and let the unpruned weight contributions be $\{+9, -9\}$. The best 1-swap is to unprune the $-1$ contribution and prune the $-9$ contribution, giving $r' = 10 + (-9) = 1$ and $L' = 1$. However, if we instead greedily remove the best $p$ in isolation, we unprune $+10$ since $(9 - 10)^2 = 1$ is minimal. We must then add one index; the best addition in isolation to the original pruned-weight-contributions $\{+10, -1\}$ is $-9$. In combination, the greedily chosen swap leads to $r' = -1 + (-9) = -10$ and $L' = 100$, \emph{worse} than the starting point. The error stems precisely from ignoring the interaction term when selecting $(p, u)$.
\subsection{The \glsshort{greedyaf} Algorithm}\label{sec:greedyaf}

Building upon the preceding observations, we present our complete algorithm. The method takes as input a weight matrix $W \in \R^{d_{out} \times d_{in}}$, the Gram matrix $G = XX^{\top} \in \R^{d_{in} \times d_{in}}$ (accumulated during calibration), and a warmstart pruning mask $M^{\text{init}} \in \{0,1\}^{d_{out} \times d_{in}}$ that already satisfies the desired sparsity constraints, e.g., obtained from Wanda \citep{Sun2023} or RIA \citep{Zhang2024}.

The algorithm enforces any sparsity pattern that operates per-row, including per-row sparsity (fixed number of zeros per row, cf. \citet{Sun2023}) and structured $N\text{:}M$ sparsity patterns (e.g., 2:4 or 4:8, \citet{Mishra2021}). All swap operations maintain the sparsity constraints throughout optimization; for $N\text{:}M$ sparsity, swaps are restricted to occur only within the same $N\text{:}M$ blocks, while for per-row sparsity, the total number of pruned weights per row remains constant. Even though each swap only changes two mask entries, the cumulative effect of multiple swaps can dramatically reduce reconstruction error compared to the initial solution.

\begin{algorithm}[h]
  \caption{\glsshort{greedyaf}: 1-Swap Pruning Optimization}
  \label{alg:swap}
  \begin{algorithmic}[1]
    \REQUIRE $W \in \R^{d_{out} \times d_{in}}$, Gram matrix $G = XX^{\top} \in \R^{d_{in} \times d_{in}}$, warmstart mask $M^{\text{init}}$, $T_{\max}$
    \ENSURE Improved pruning mask $M$

    \STATE $M \leftarrow M^{\text{init}}$ \hfill \textit{Warmstart initialization}

    \FOR{$i = 1$ to $d_{out}$}
      \STATE $w \leftarrow W_{i,:}$, $m \leftarrow M_{i,:}$ 
      \STATE $\mathcal P \leftarrow \{j : m_j = 0\}$, $\mathcal U \leftarrow \{j : m_j = 1\}$ 

      \STATE $c \leftarrow G \cdot ((\mathbbm{1} - m) \odot w)$ 

      \FOR{$t = 1$ to $T_{\max}$}
        \STATE $(p^*, u^*) \gets \arg\min_{(p,u)} \; \Delta L_{u,p}$ \hfill \textit{Best swap} 

        \IF{$\Delta L_{u^*,p^*} < 0$}
          \STATE $m_{p^*} \leftarrow 1$, $m_{u^*} \leftarrow 0$ \hfill \textit{Perform swap}
          \STATE $\mathcal P \leftarrow (\mathcal P \setminus \{p^*\}) \cup \{u^*\}$, $\mathcal U \leftarrow (\mathcal U \setminus \{u^*\}) \cup \{p^*\}$
          \STATE $c \leftarrow c + w_{u^*} G_{:,u^*} - w_{p^*} G_{:,p^*}$ 
        \ELSE
          \STATE \textbf{break} \hfill \textit{Local optimum reached}
        \ENDIF
      \ENDFOR
      \STATE $M_{i,:} \leftarrow m$ \hfill \textit{Store optimized row}
    \ENDFOR
  \end{algorithmic}
\end{algorithm}

We explain the main phases of the algorithm:

\textbf{Preparation:} We initialize with the warmstart mask $M^{\text{init}}$. The Gram matrix $G$ is precomputed once per layer by accumulating $G = \sum_b X_{:,b} X_{:,b}^{\top}$ during the calibration forward pass.

\textbf{Row processing (Lines 2-5):} For each row $i$, we extract weights $w$ and current mask $m$, define pruned and unpruned index sets $\mathcal{P}$ and $\mathcal{U}$, and compute the initial correlation vector $c = G \cdot ((\mathbbm{1} - m) \odot w)$.

\textbf{1-Swap optimization (Lines 6-15):} We iteratively find the swap $(p^*, u^*)$ minimizing $\Delta L_{u,p}$ (cf. \autoref{eq:swap_cost}) among feasible pairs, evaluating each candidate using only lookups. If $\Delta L_{u^*,p^*} < 0$, we accept the swap and update the correlation vector via \autoref{eq:c_update}; otherwise we terminate. At all times, the swaps are appropriately constrained: per-row sparsity allows any swap maintaining $|\mathcal{P}|$ constant, while $N\text{:}M$ sparsity restricts swaps to within the same $N\text{:}M$ blocks.

The algorithm has complexity $\mathcal{O}(d_{out} \cdot T_{\max} \cdot (|\mathcal{P}| \cdot |\mathcal{U}| + d_{in}))$ per layer, where $T_{\max}$ is the maximum number of swap iterations per row. The $|\mathcal{P}| \cdot |\mathcal{U}|$ term comes from evaluating all candidate swaps (each in $\mathcal{O}(1)$ time via \autoref{eq:swap_cost}), and the $d_{in}$ term from the correlation vector update (\autoref{eq:c_update}).

\paragraph{Convergence.} Intuitively, since each accepted swap strictly decreases the per-row loss $L$ and $L \geq 0$, the algorithm must terminate at a \emph{1-swap local optimum}, i.e., a mask from which no single swap can reduce the loss. We state this simple observation below and defer the formal statement and proof to the appendix.

\begin{proposition}[Convergence, informal]\label{prop:convergence}
Suppose \cref{alg:swap} terminates when no swap improves the loss by more than $\varepsilon \geq 0$. Then, starting from initial per-row loss $L_0$, it performs at most $\mathcal{O}(L_0 / \varepsilon)$ swaps to reach a $\varepsilon$-1-swap local optimum.
\end{proposition}

In practice, several factors further reduce the runtime. First, we find that even $T_{\max}=1$ or $T_{\max}=2$ can drastically reduce the local pruning error; values around $T_{\max} = 25$ often suffice to significantly lower model perplexity, with diminishing returns beyond $T_{\max} = 100$. Second, row-wise processing can be batched and vectorized, enabling parallel swap cost computations and mask updates, and rows can be distributed across GPUs if needed. Third, the Gram matrix $G$ is computed once per layer and shared across all rows, and several summands of \autoref{eq:swap_cost} can be similarly precomputed once per layer.

\section{Experimental Results}\label{sec:experimental-results}
We outline our general experimental approach, detailing datasets, architectures, and metrics. Our code is publicly available at \href{https://github.com/ZIB-IOL/SparseSwaps}{github.com/ZIB-IOL/SparseSwaps}. We focus on language modeling within \gls{nlp}. We use pretrained models from HuggingFace \citep{Wolf2020}, specifically \textsc{LLaMA-3.1-8B} \citep{grattafioriLlama3Herd2024}, \textsc{Gemma-2-9B} \citep{riviereGemma2Improving2024}, \textsc{Yi-1.5-9B} \citep{youngYiOpenFoundation2025}, \textsc{DeepSeek-7B-base} \citep{biDeepSeekLLMScaling2024}, and \textsc{Qwen2.5-7B} \citep{yangQwen25TechnicalReport2025}. For calibration, we randomly draw sequences of 2048 tokens from the \emph{C4} dataset \citep{Raffel2020a}. For validation, we similarly pick 100 sequences from the validation split. The model performance is assessed via perplexity on the \emph{WikiText} dataset \citep{Merity2016} and zero-shot accuracy on the EleutherAI evaluation set \citep{Gao2023}. Following \citet{Sun2023}, we prune all linear layers, excluding the embedding and final linear head, with uniform sparsity allocation across layers. We provide experiments for unstructured and semi-structured sparsity patterns \citep{Mishra2021}. We use multiple random seeds throughout our experiments.

\subsection{Mask Refinement at Scale}
We begin by verifying the effectiveness of \glsshort{greedyaf}. We make the following observations:

\begin{table*}[t]
  \caption{Perplexity ($\downarrow$, lower is better) and zero-shot accuracy ($\uparrow$, higher is better) comparison on WikiText and EleutherAI evaluation set. We report \glsshort{dsnot} and \glsshort{greedyaf} refinement with Wanda and RIA warmstart for unstructured 60\% sparsity and semi-structured 2:4 sparsity. Best values are highlighted in \textbf{bold}. We omit standard deviations for legibility.}
  \label{tab:accuracy_reconstruction_comparison}
  \centering
  \resizebox{\textwidth}{!}{%
  \begin{tabular}{l c c c c c c c c c c c c}
  \toprule
  & & \multicolumn{5}{c}{\textbf{Perplexity $\downarrow$}} & & \multicolumn{5}{c}{\textbf{Accuracy $\uparrow$}} \\
  \cmidrule{3-7} \cmidrule{9-13}
  \textbf{Method} & \textbf{Sparsity} & \textbf{\textsc{LLaMA-3.1}} & \textbf{\textsc{Gemma-2}} & \textbf{\textsc{Yi-1.5}} & \textbf{\textsc{DeepSeek}} & \textbf{\textsc{Qwen2.5}} & & \textbf{\textsc{LLaMA-3.1}} & \textbf{\textsc{Gemma-2}} & \textbf{\textsc{Yi-1.5}} & \textbf{\textsc{DeepSeek}} & \textbf{\textsc{Qwen2.5}} \\
  & & \textbf{8B} & \textbf{9B} & \textbf{9B} & \textbf{7B} & \textbf{7B} & & \textbf{8B} & \textbf{9B} & \textbf{9B} & \textbf{7B} & \textbf{7B} \\
  \midrule
  Wanda & 60\% & 21.94 & 16.74 & 11.40 & 11.41 & 13.75 & & 48.18\% & 63.39\% & 53.59\% & 50.74\% & 59.26\% \\
  \texttt{+} \glsshort{dsnot} & 60\% & 21.94 & 16.69 & 11.38 & 11.40 & 13.75 & & 48.18\% & 63.49\% & 53.79\% & 50.75\% & 59.26\% \\
  \textbf{\texttt{+} \glsshort{greedyaf}} & 60\% & \textbf{19.75} & \textbf{16.01} & \textbf{10.07} & \textbf{10.93} & \textbf{13.16} & & \textbf{50.78\%} & \textbf{63.84\%} & \textbf{54.84\%} & \textbf{51.02\%} & \textbf{60.15\%} \\
  \midrule
  RIA & 60\% & 19.73 & 16.19 & 10.73 & 11.80 & 12.63 & & 49.56\% & 64.37\% & 52.81\% & 50.92\% & 59.84\% \\
  \texttt{+} \glsshort{dsnot} & 60\% & 19.73 & 16.22 & 10.73 & 11.80 & 12.63 & & 49.56\% & \textbf{64.43\%} & 52.96\% & 50.83\% & 59.81\% \\
  \textbf{\texttt{+} \glsshort{greedyaf}} & 60\% & \textbf{18.47} & \textbf{15.44} & \textbf{9.98} & \textbf{10.79} & \textbf{12.47} & & \textbf{51.02\%} & 64.32\% & \textbf{54.45\%} & \textbf{51.47\%} & \textbf{61.22\%} \\
  \midrule
  Wanda & 2:4 & 24.82 & 17.45 & 11.76 & 11.77 & 14.53 & & 46.80\% & 63.73\% & \textbf{52.58\%} & \textbf{51.02\%} & 59.52\% \\
  \texttt{+} \glsshort{dsnot} & 2:4 & 22.79 & 16.79 & 10.84 & 11.70 & 14.40 & & 47.01\% & 63.66\% & 52.16\% & 50.78\% & 59.09\% \\
  \textbf{\texttt{+} \glsshort{greedyaf}} & 2:4 & \textbf{20.17} & \textbf{16.30} & \textbf{10.73} & \textbf{11.70} & \textbf{13.95} & & \textbf{48.83\%} & \textbf{64.70\%} & 52.43\% & 50.36\% & \textbf{59.92\%} \\
  \midrule
  RIA & 2:4 & 23.96 & 16.88 & 11.29 & 12.03 & 13.58 & & 47.87\% & 63.87\% & \textbf{52.68\%} & 51.22\% & 58.66\% \\
  \texttt{+} \glsshort{dsnot} & 2:4 & 24.26 & 16.82 & 10.57 & 12.03 & 13.85 & & 47.13\% & 64.17\% & 51.36\% & 49.86\% & 59.72\% \\
  \textbf{\texttt{+} \glsshort{greedyaf}} & 2:4 & \textbf{20.90} & \textbf{16.33} & \textbf{10.50} & \textbf{11.80} & \textbf{13.28} & & \textbf{49.90\%} & \textbf{64.60\%} & 52.30\% & \textbf{51.46\%} & \textbf{60.31\%} \\
  \bottomrule
  \end{tabular}
  }
\end{table*}

\textbf{\glsshort{greedyaf} consistently improves state-of-the-art methods.} \autoref{tab:accuracy_reconstruction_comparison} summarizes the main results and reports perplexity (upper half, lower is better) and zero-shot accuracy (lower half, higher is better) for warmstart masks (Wanda, RIA) as well as their refinements using \glsshort{dsnot} and \glsshort{greedyaf}. For both 60\% unstructured and 2:4 semi-structured sparsity, \glsshort{greedyaf} (with 100 1-swap iterations) consistently reduces perplexity and improves zero-shot accuracy over Wanda and RIA warmstart masks. While \glsshort{dsnot} similarly yields improvements, it falls short of \glsshort{greedyaf}. Note that we left the pruning criterion of \glsshort{dsnot}, which partially uses the Wanda saliency, unchanged, even when using RIA warmstart. For unstructured RIA, we report results when enforcing a per-row sparsity constraint; while RIA yields good (and slightly better) results when enforcing truly unstructured sparsity, we decided to include the results for the per-row setting as this allows direct refinement of the mask with \glsshort{greedyaf} and \glsshort{dsnot}.

\textbf{\glsshort{greedyaf} successfully optimizes the per-layer pruning loss.} \autoref{fig:layer_error} shows the per-layer reductions in local pruning error relative to a Wanda warmstart, grouping layers by their corresponding Transformer block of \textsc{LLaMA-3.1-8B}. We observe drastic improvements of close to 70\% compared to Wanda, demonstrating that \glsshort{greedyaf} is able to successfully optimize the local loss. The \texttt{attn.o\_proj} seems to consistently benefit the most across blocks, with reductions of the objective in \autoref{eq:mask_selection_problem} ranging between 40\% and 60\%.

\begin{table}[h]
  \caption{Perplexity ($\downarrow$, lower is better) comparison on WikiText. We report \glsshort{greedyaf} refinement with magnitude warmstart for 
  50\% and 60\% sparsity. Best values are highlighted in \textbf{bold}. We omit standard deviations for legibility.}
  \label{tab:accuracy_reconstruction_comparison_magnitude}
  \centering
  \resizebox{\columnwidth}{!}{%
  \begin{tabular}{lc c c c}
  \toprule
  \footnotesize{\textbf{Perplexity $\downarrow$}} & & \textbf{\textsc{LLaMA-3.1}} & \textbf{\textsc{Gemma-2}} & \textbf{\textsc{DeepSeek}} \\
  \cmidrule{3-3} \cmidrule{4-4} \cmidrule{5-5}
      \textbf{Method} & Sparsity & 8B & 9B & 7B \\
  \midrule
  Magnitude & 50\% & 68.89 & 31.87 & 25.05  \\
  \textbf{\texttt{+} \glsshort{greedyaf}} & 50\% & 52.26 & 19.11 & 16.23  \\
  \midrule
  Magnitude & 60\% & 3486.26 & 184.52 & 330.07  \\
  \textbf{\texttt{+} \glsshort{greedyaf}} & 60\% & 264.92 & 60.04 & 80.24  \\
  \bottomrule
  \end{tabular}
  }
  \end{table}

\textbf{Large local error reductions do not always imply reduced perplexity.} From \autoref{tab:accuracy_reconstruction_comparison} we observe substantial perplexity gains. Surprisingly, when the model quality is less affected by pruning (e.g., at 50\% sparsity where Wanda performs well), \glsshort{greedyaf} yields limited perplexity gains despite significant local error reductions: \autoref{tab:abl_max_passes} reports perplexity and average relative error reduction (\%) versus the number of 1-swap iterations. Zero iterations correspond to the Wanda warmstart; one or more iterations correspond to \glsshort{greedyaf} from Wanda. At 50\% sparsity, a single 1-swap iteration lowers relative error by 6.34\%, and 100 iterations by close to 40\%, yet perplexity does not improve, but rather slightly increases. This suggests further reducing local error can overfit the calibration data and may not translate to better perplexity, although we note that the perplexity increase is relatively small. These results emphasize that while the reduction of local error is a useful proxy for perplexity reduction when pruning has a higher negative impact on the model, the local error of \autoref{eq:mask_selection_problem} remains only an approximation to the reconstruction error of the entire model. \autoref{tab:accuracy_reconstruction_comparison_magnitude} shows more drastic improvements when using magnitude pruning, further highlighting that the impact of \glsshort{greedyaf} is most pronounced when model degradation is high.

\begin{table}[h]
  \caption{\textsc{LLaMA-3.1-8B}: Mean relative reduction in pruning error ($\uparrow$) and perplexity ($\downarrow$) versus number of 1-swap iterations for 50\% and 60\% unstructured sparsity using Wanda warmstart.}
  \label{tab:abl_max_passes}
  \centering
  \resizebox{\columnwidth}{!}{%
  \begin{tabular}{c c c c c c c c c c}
  & & \multicolumn{8}{c}{\textbf{Number of 1-swap iterations}} \\
  \toprule
  \textbf{Sparsity} & \textbf{Metric} & 0 & 1 & 2 & 5 & 10 & 25 & 50 & 100 \\
  \midrule
  \multirow{2}{*}{\textbf{50\%}} & Error reduction (\%) & 0.00 & 6.34 & 8.77 & 12.51 & 16.38 & 23.52 & 30.04 & 36.48\\
  & Perplexity & 10.13 & 10.31 & 10.40 & 10.41 & 10.39 & 10.38 & 10.27 & 10.30 \\
  \midrule
  \multirow{2}{*}{\textbf{60\%}} & Error reduction (\%) & 0.00 & 8.04 & 11.04 & 15.34 & 19.64 & 26.92 & 33.58 & 39.99\\
  & Perplexity & 21.52 & 21.26 & 21.51 & 21.17 & 21.01 & 20.38 & 19.74 & 18.96 \\
  \bottomrule
  \end{tabular}
  }
\end{table}

\textbf{\glsshort{greedyaf} is robust to warmstart quality.} \autoref{tab:warmstart_sensitivity} shows that weaker warmstarts yield larger error reductions (58--63\% from magnitude vs.\ 37--43\% from Wanda), demonstrating that \glsshort{greedyaf} is not sensitive to warmstart quality and can recover from poor initializations.

\begin{table}[h]
  \caption{Average relative reduction in local pruning error (\%) for different warmstart methods at 60\% sparsity. Weaker warmstarts yield larger reductions.}
  \label{tab:warmstart_sensitivity}
  \centering
  \resizebox{\columnwidth}{!}{%
  \begin{tabular}{l c c c}
  \toprule
  \textbf{Warmstart} & \textbf{\textsc{LLaMA-3.1-8B}} & \textbf{\textsc{Gemma-2-9B}} & \textbf{\textsc{DeepSeek-7B}} \\
  \midrule
  Magnitude & 62.69\% & 58.13\% & 58.10\% \\
  Wanda & 43.29\% & 40.45\% & 36.96\% \\
  \bottomrule
  \end{tabular}
  }
\end{table}

\subsection{Efficiency and Hyperparameter Ablations}
\textbf{Resource requirements.}
\glsshort{greedyaf} is more resource-intensive than \glsshort{dsnot} and, as a drop-in refinement, requires at least the resources of the chosen warmstart method. Beyond that, \glsshort{greedyaf} needs memory to store the Gram matrix $G \in \R^{d_{in} \times d_{in}}$ (once per layer) and the correlation vector $c \in \R^{d_{in}}$ (per row), as well as the compute to perform the 1-swaps; see the preceding section for the theoretical complexity. While we have argued in the introduction that the additional compute can be justified when amortized over many \gls{llm} inference requests, we note that the overhead grows only linearly with the number of 1-swap iterations $T_{\max}$. \autoref{tab:abl_max_passes} shows that few iterations already yield substantial gains in both perplexity and local error reduction, especially at higher sparsity.

\begin{table}[h]
  \caption{Wall-clock time in minutes for applying \glsshort{greedyaf} to \textsc{LLaMA-3.1-8B} at 60\% sparsity on a single H100 GPU. The $T_{\max}=0$ baseline includes calibration data sampling, Wanda pruning, Gram matrix computation, and evaluation.}
  \label{tab:runtime}
  \centering
  \resizebox{\columnwidth}{!}{%
  \begin{tabular}{l c c c c c c}
  \toprule
  $\mathbf{T_{\max}}$ & 0 & 1 & 2 & 5 & 10 & 25 \\
  \midrule
  Wall-clock time & 8m & 10m & 12m & 17m & 26m & 52m \\
  \bottomrule
  \end{tabular}
  }
\end{table}

\autoref{tab:runtime} further reports wall-clock times for pruning \textsc{LLaMA-3.1-8B} to 60\% sparsity on a single H100 GPU. The $T_{\max}=0$ baseline includes calibration data sampling, Wanda pruning, Gram matrix computation, and evaluation; each additional iteration of \glsshort{greedyaf} adds a relatively small overhead. For comparison, Wanda and SparseGPT take approximately 4 and 10 minutes, respectively. We note that our implementation can be further optimized and that the algorithm is fully parallelizable across rows.

\textbf{Effect of the number of reconstruction samples.} \autoref{fig:perplexity_vs_reconstruct_n_samples} in the appendix shows the perplexity versus the number of reconstruction samples for 50\% and 60\% unstructured sparsity when using Wanda as well as \glsshort{greedyaf} with a Wanda warmstart. We observe that the perplexity decreases drastically when using more samples, which leads to \glsshort{greedyaf} slightly outperforming Wanda for 50\% sparsity, despite its advantage typically being larger at higher sparsity. We emphasize that the number of reconstruction samples does not affect \glsshort{greedyaf}' swap evaluation efficiency: the Gram matrix $G = XX^{\top}$ has fixed size $d_{in} \times d_{in}$ regardless of $B$.

\section{Conclusion}\label{sec:conclusion}
We revisited the mask selection problem for post-training pruning and showed that it can be made substantially more tractable, even at \gls{llm} scale. We observed that row decoupling via equal per-row sparsity yields independent subproblems, and that individual 1-swaps can be evaluated in $\mathcal{O}(1)$ time using the Gram matrix $G = XX^{\top}$. This enables tractable optimization of the true row-wise quadratic loss on GPUs. The resulting method, \glsshort{greedyaf}, is warmstart-agnostic, nearly hyperparameter-free, and scalable. It consistently reduces per-layer pruning error and improves perplexity and zero-shot accuracy across modern \gls{gpt} architectures.

Our work is not without limitations. While per-row sparsity is not necessarily detrimental for \glspl{llm}, our approach is restricted to that setting and only partially adapts to truly unstructured sparsity; in its current form, the algorithm can handle unstructured sparsity but cannot reallocate sparsity levels across rows. A reallocation of sparsity between individual rows might pose an interesting direction for further research. Further, the focus of our work is to demonstrate that exact mask refinement is realizable at \gls{llm} scale and can improve existing mask-selection methods. Our objective is to refine pruning masks post-hoc. There exists a variety of methods that optimize both the mask as well as the remaining, non-pruned weights jointly. In contrast, \glsshort{greedyaf} focuses on mask optimization and is complementary to such approaches, which can in principle be combined with \glsshort{greedyaf} to refine intermediate pruning steps. We leave such extensions for future work.

\section*{Acknowledgments}
This research was partially supported by the DFG Cluster of Excellence MATH+ (EXC-2046/2, project id 390685689) funded by the Deutsche Forschungsgemeinschaft (DFG) as well as by the German Federal Ministry of Research, Technology and Space (research campus Modal, fund number 05M14ZAM, 05M20ZBM) and the VDI/VDE Innovation + Technik GmbH (fund number 16IS23025B).

\bibliography{max_references,max_zotero_references,deborah_references}
\bibliographystyle{icml2026}

\newpage
\appendix
\onecolumn
\section{Appendix}

A different (but in practice slightly less efficient) perspective on the reduction of the computational complexity through the Gram matrix is through the \emph{unitary invariance} of the Frobenius norm used in our pruning objective: for any matrix $A$ and unitary matrix $U$ (i.e., $U^{-1} = U^{\top}$), we have $\norm{AU}_F = \norm{A}_F$. This property enables significant computational savings through \gls{svd} compression. Precisely, let $X = U\Sigma V^{\top}$ be the \gls{svd} of calibration data $X \in \mathbb{R}^{d_{in} \times B}$. Since $B>d_{in}$, we can write $\Sigma=[\Sigma' \mid 0]$ with $\Sigma' \in \R^{d_{in} \times d_{in}}$ containing the singular values on its diagonal. The compressed representation is simply $X' = U\Sigma' \in \R^{d_{in} \times d_{in}}$. Letting $w_p = w - m \odot w$ for brevity, the key insight is that pruning decisions remain equivalent under this compression:
  \begin{align*}
      \norm{w_pX}_F^2 
      &= \norm{w_pU\Sigma V^{\top}}_F^2 
      = \norm{w_pU\Sigma}_F^2\\ 
      &= \norm{w_pU[\Sigma' \mid 0]}_F^2 
      = \norm{w_pU\Sigma'}_F^2 
      = \norm{w_pX'}_F^2,
  \end{align*}
  where we used unitary invariance w.r.t. $V$ and that the zero columns do not contribute to the Frobenius norm. Equivalently, we have
  \begin{equation*}
    X'X'^{\top} = U\Sigma'\Sigma'^{\top}U^{\top} = U\Sigma\Sigma^{\top}U^{\top} = XX^{\top} = G,
  \end{equation*}
  since $\Sigma\Sigma^{\top} = \Sigma'\Sigma'^{\top}$ (the zero columns of $\Sigma$ do not contribute). Since all subsequent operations depend solely on $G$, we accumulate $G$ directly during calibration and avoid the \gls{svd} entirely.

\subsection{Convergence to a Local Optimum}\label{app:convergence}

We formalize the convergence guarantee for \glsshort{greedyaf}. Recall that for a single row with pruned index set $\mathcal{P}$ and unpruned set $\mathcal{U}$, the per-row loss is $L = \|r\|_2^2$ where $r = \sum_{j \in \mathcal{P}} w_j \phi_j$. A 1-swap exchanges one index between $\mathcal{P}$ and $\mathcal{U}$, preserving the sparsity level.

\begin{definition}[$\varepsilon$-1-swap local optimum]
A mask $m$ (equivalently, its pruned set $\mathcal{P}$) is an \emph{$\varepsilon$-1-swap local optimum} if
\begin{equation*}
\min_{u \in \mathcal{U}, p \in \mathcal{P}} \Delta L_{u,p} \geq -\varepsilon,
\end{equation*}
i.e., no single swap can reduce the per-row loss $L$ by more than $\varepsilon$.
\end{definition}

\begin{proposition}[Convergence]\label{prop:convergence_formal}
Let $m_0$ be the initial mask with per-row loss $L_0$. Suppose \glsshort{greedyaf} terminates when the best swap satisfies $\Delta L_{u^*,p^*} \geq -\varepsilon$ for some $\varepsilon \geq 0$. Then \glsshort{greedyaf} performs at most $T \leq \lceil L_0 / \varepsilon \rceil$ swap operations before reaching an $\varepsilon$-1-swap local optimum.
\end{proposition}

\begin{proof}
At any iteration $t$ that does not terminate, the algorithm accepts a swap with $\Delta L_{u^*,p^*} < -\varepsilon$ by definition of the stopping rule. Since $L_{t+1} = L_t + \Delta L_{u^*,p^*}$, this yields the strict decrease $L_{t+1} < L_t - \varepsilon$. Unrolling over $T$ accepted swaps gives $L_T < L_0 - T\varepsilon$. Since $L$ is a squared Euclidean norm, we have $L_T \geq 0$, which implies $T < L_0 / \varepsilon$, and hence $T \leq \lceil L_0 / \varepsilon \rceil$. At termination, we have $\min_{(u,p)} \Delta L_{u,p} \geq -\varepsilon$, so the final mask is an $\varepsilon$-1-swap local optimum.
\end{proof}

\subsection{Further Results}
\begin{figure}[h]
      \centering
      \begin{minipage}{0.48\linewidth}
        \centering
        \includegraphics[width=\linewidth]{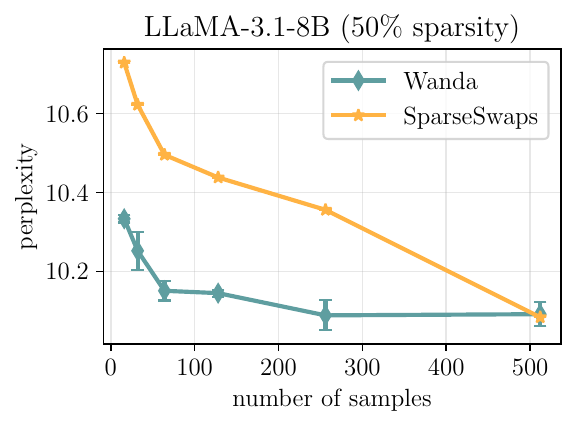}
        \subcaption{50\% unstructured sparsity}
        \label{fig:perplexity_vs_reconstruct_n_samples_50}
      \end{minipage}
      \hfill
      \begin{minipage}{0.48\linewidth}
        \centering
        \includegraphics[width=\linewidth]{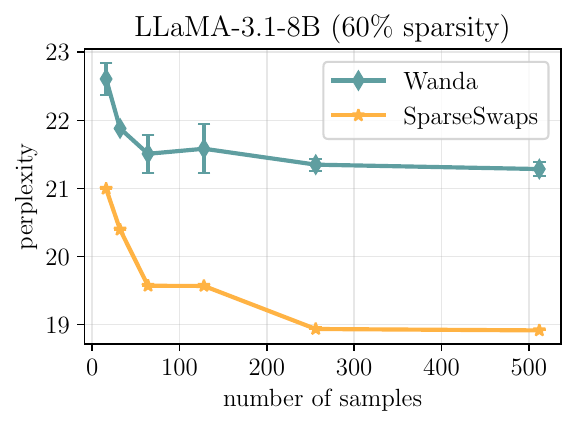}
        \subcaption{60\% unstructured sparsity}
        \label{fig:perplexity_vs_reconstruct_n_samples_60}
      \end{minipage}
      \caption{Perplexity versus the number of reconstruction samples for unstructured sparsity using Wanda warmstart.}
      \label{fig:perplexity_vs_reconstruct_n_samples}
    \end{figure}

\autoref{fig:perplexity_vs_reconstruct_n_samples} shows the perplexity versus the number of reconstruction samples for 50\% and 60\% unstructured sparsity when using Wanda as well as \glsshort{greedyaf} with a Wanda warmstart. We further ran an ablation study comparing Wanda and \glsshort{greedyaf} when varying the sequence length while keeping the number of calibration samples fixed at 128. Both methods were equally impacted by the sequence length, with the overall impact being rather minor in terms of final perplexity.

\end{document}